\documentclass[runningheads]{llncs}
\usepackage{graphicx}
\usepackage{comment}
\usepackage{amsmath,amssymb} 
\usepackage{color}
\usepackage{appendix}

\usepackage{subfigure}
\usepackage{amsfonts,bm}
\usepackage{algorithm}
\usepackage{algorithmic}
\usepackage{multirow}
\usepackage{graphicx}
\usepackage{enumitem}
\usepackage{verbatimbox}

\usepackage[colorlinks,linkcolor=blue]{hyperref}
\def\st{{\rm{s.t.}}}

\newfloat{figtab}{htb}{fgtb}
\makeatletter
  \newcommand\figcaption{\def\@captype{figure}\caption}
  \newcommand\tabcaption{\def\@captype{table}\caption}
\makeatother
\usepackage{multirow}
\usepackage{booktabs}

\DeclareMathOperator*{\argmax}{arg\,max}
\DeclareMathOperator*{\argmin}{arg\,min}

\def\st{{\rm{s.t.}}}

\begin{document}
\pagestyle{headings}
\mainmatter
\def\ECCVSubNumber{3412}  

\title{Motion-Excited Sampler: Video Adversarial Attack with Sparked Prior} 

\titlerunning{Motion-Excited Sampler: Video Adversarial Attack with Sparked Prior}
\author{Hu Zhang\inst{1} \and
Linchao Zhu\inst{1} \and
Yi Zhu\inst{2} \and
Yi Yang\inst{1}}
\authorrunning{Hu Zhang, Linchao Zhu, Yi Zhu and Yi Yang}
%
\institute{ReLER, University of Technology Sydney, NSW \and
Amazon Web Services \\
\email{Hu.Zhang-1@student.uts.edu.au; zhulinchao7@gmail.com; }\\
\email{yzaws@amazon.com; Yi.Yang@uts.edu.au}}
\maketitle

\begin{abstract}
Deep neural networks are known to be susceptible to adversarial noise, which is tiny and imperceptible perturbation. Most of previous works on adversarial attack mainly focus on image models, while the vulnerability of video models is less explored. In this paper, we aim to attack video models by utilizing intrinsic movement pattern and regional relative motion among video frames. We propose an effective motion-excited sampler to obtain motion-aware noise prior, which we term as sparked prior. Our sparked prior underlines frame correlations and utilizes video dynamics via relative motion. By using the sparked prior in gradient estimation, we can successfully attack a variety of video classification models with fewer number of queries. Extensive experimental results on four benchmark datasets validate the efficacy of our proposed method. The code can be found here: \href{https://github.com/xiaofanustc/ME-Sampler}{https://github.com/xiaofanustc/ME-Sampler}.
\end{abstract}

\section{Introduction}
\label{sec:introduction}

Despite the superior performance achieved in a variety of computer vision tasks, i.e., image classification \cite{He2016resnet}, object detection \cite{renNIPS15fasterrcnn}, segmentation \cite{he2017maskrcnn,chen2018deeplabv3plus}, Deep Neural Networks (DNNs) are shown to be susceptible to adversarial attacks that a well-trained DNN classifier may make severe mistakes due to a single invisible perturbation on a benign input and suffer dramatic performance degradation. To investigate the vulnerability and robustness of DNNs, many effective attack methods have been proposed on image models. They either consider a white-box attack setting where the adversary can always get the full access to the model including exact gradients of given input, or a black-box one, in which the structure and parameters of the model are blocked that the attacker can only access the ($input, output$) pair through queries.
\begin{figure}[t]
\centering
\scalebox{1.0}{
\subfigure[]{
\label{fig:(a)}
   \includegraphics[width=0.38\linewidth]{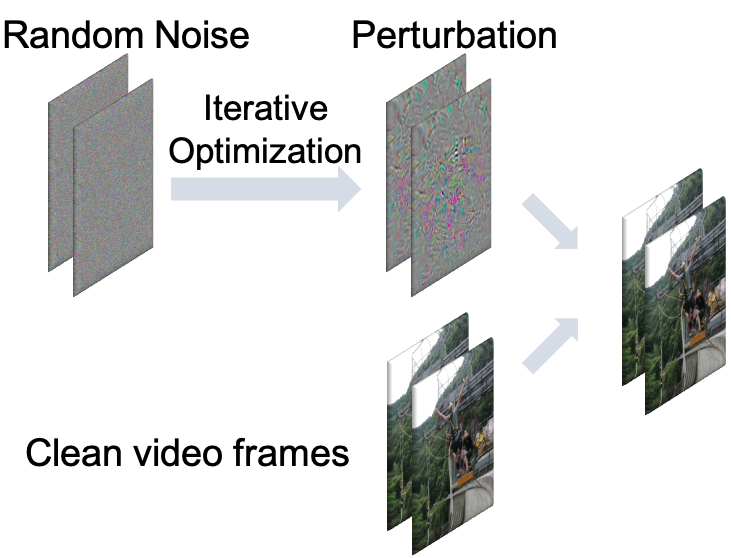}}
\subfigure[]{
\label{fig:(b)}
   \includegraphics[width=0.38\linewidth]{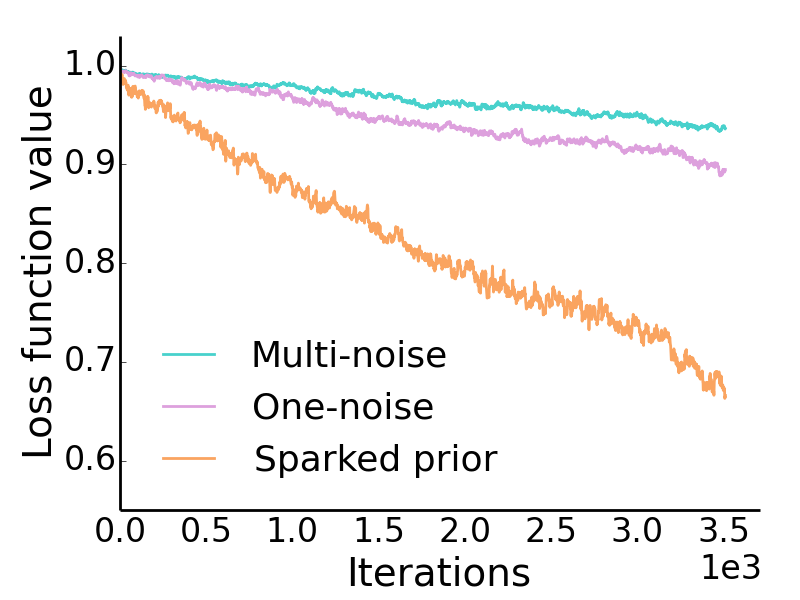}}}
   \caption{(a) A pipeline of generating adversarial examples to attack a video model. (b) Loss curve comparison: i) Multi-noise: sample noise prior individually for each frame; ii) One-noise: sample one noise prior for all frames; iii) Sparked prior (ours): sample one noise prior for all frames and sparked by motion information. Loss is computed in attacking an I3D model on Kinetics-400 dataset. The lower loss indicates the better attacking performance. Our proposed sparked prior clearly outperforms (i) and (ii) in terms of attacking video models. The figure is best viewed in color.}
\label{fig1: kinetics loss curve}
\end{figure}

DNNs have also been widely applied in video tasks, such as video action recognition \cite{Wang2016TSN,kay2017kinetics}, video object detection \cite{wu2019selsa}, video segmentation \cite{wang2019fast}, video inpainting \cite{kim2019deep} etc. However, limited work has been done on attacking video models. A standard pipeline of video adversarial attack is shown in Fig.~\ref{fig1: kinetics loss curve}(a). Specially designed perturbations are estimated from prior, which normally is random noise, and imposed on the clean video frames to generate the adversarial examples. The goal is using the adversarial examples to trick the model into giving a wrong prediction. Most literature~\cite{wei2018sparse,li2018adversarial} focuses on the white-box attack setting and simply transfer the methods used in image domain to video domain. Recently \cite{jiang2019black} proposes a black-box attack method, which simply decodes a video into frames and transfers gradients from a pretrained image model for each frame. All the aforementioned methods ignore the intrinsic difference between images and videos, e.g., the extra temporal dimension.
This naturally leads to a question: should we use motion information in video adversarial attack?

In this work, we propose to use motion information for black-box attack on video models. In each optimization step, instead of directly using random noise as prior, we first generate a motion map (i.e., motion vector or optical flow) between frames and construct a motion-excited sampler. The random noise will then be selected by the sampler to obtain motion-aware noise for gradient estimation, which we term as sparked prior. In the end, we feed the original video frames and the sparked prior to gradient estimator and use the estimated gradients to iteratively update the video. To show the effectiveness of using motion information, we perform a proof-of-concept comparison to two baselines. One is initializing noises separately for each frame extending~\cite{ilyas2018prior} (multi-noise), the other is initializing one noise and replicating it across all video frames (one-noise).
We use the training loss curve to reflect the effectiveness of video attack methods, in which the loss measures the distance to a fake label. As we can see in Fig. \ref{fig1: kinetics loss curve}(b), the loss of our proposed method drops significantly faster (orange curve) than one-noise and multi-noise method. This indicates that our method takes fewer queries to successfully attack a video model. This answers our previous question that we should use motion information in video adversarial attack.  Our main contributions can be summarized as follows: 

\begin{itemize}
\item We find that simply transferring attack methods on image models to video models is less effective. Motion plays a key role in video attacking.
\item We propose a motion-excited sampler to obtain sparked prior, which leads to more effective gradient estimation for faster adversarial optimization. 
\item We perform thorough experiments on four video action recognition datasets against two kinds of models and show the efficacy of our proposed algorithm.
\end{itemize}

\section{Related Work}
\label{sec:related_work}
\noindent\textbf{Adversarial Attack.}
Adversarial examples have been well studied on image models. \cite{szegedy2013intriguing} first shows that an adversarial sample, computed by imposing small noise on the original image, could lead to a wrong prediction.
By defining a number of new losses, \cite{carlini2017towards} demonstrates that previous defense methods do not significantly increase the robustness of neural networks. \cite{papernot2017practical} first studies the black-box attack in image model by leveraging the transferability of adversarial examples, however, their success rate is rather limited. 
\cite{ilyas2018black} extends Natural Evolutionary Strategies~(NES) to do gradient estimation and \cite{ilyas2018prior} proposes to use time and data-dependent priors to reduce queries in black-box image attack. More recently, \cite{Du2020Query-efficient} proposes a meta-based method for query-efficient attack on image models.

However, limited work have been done on attacking video models. In terms of white-box attack, \cite{wei2018sparse} proposes to investigate the sparsity of adversarial perturbations and their propagation across video frames. \cite{li2018adversarial} leverages a Generative Adversarial Network~(GAN) to account for temporal correlations and generate adversarial samples for a real-time video classification system. \cite{inkawhich2018adversarial} focuses on attacking the motion stream in a two-stream video classifier by extending \cite{goodfellow2014explaining}. \cite{chen2019appending} proposes to append a few dummy frames to attack different networks by optimizing specially designed loss.
The first black-box video attack method is proposed in \cite{jiang2019black}, where they utilize the ImageNet pretrained models to generate a tentative gradient for each video frame and use NES to rectify it. More recently,  
\cite{wei2019heuristic} and \cite{yan2020sparse} focus on sparse perturbations only on the selected frames and regions, instead of the whole video.

Our work is different from \cite{jiang2019black} because we leverage the motion information directly in generating adversarial videos. We do not utilize the ImageNet pretrained models to generate gradient for each video frame. Our work is also different from \cite{wei2019heuristic,yan2020sparse} in terms of both problem setting and evaluation metric. We follow the setting of \cite{jiang2019black} to treat the entire video as integrity, instead of attacking the video model from the perspective of frame difference in a sparse attack setting. We use attack success rate and consumed queries to evaluate our method instead of mean absolute perturbation. By using our proposed motion-aware sparked prior, we can successfully attack a number of video classification models using much fewer queries.

\noindent\textbf{Video Action Recognition.}
Recent methods for video action recognition can be categorized into two families depending on how they reason motion information, i.e., 3D Convolutional Neural Networks (CNNs)~\cite{ji20123d,tran2015learning,carreira2017quo,wang2018nonlocal,Feichtenhofer2019slowfast,zhu2020faster} and two-stream networks~\cite{simonyan2014two,Feichtenhofer2016twofusion,Wang2016TSN,Zhu2018Hidden,lin2019tsm}.
3D CNNs simply extend 2D filters to 3D filters in order to learn spatio-temporal representations directly from videos. Since early 3D models~\cite{ji20123d,tran2015learning} are hard to train, many follow-up work have been proposed~\cite{carreira2017quo,qiu2017p3d,tran2018r21d,Feichtenhofer2019slowfast}.
Two-stream methods \cite{simonyan2014two} train two separate networks, a spatial stream given input of RGB images and a temporal stream given input of stacked optical flow images. An early \cite{Feichtenhofer2016twofusion} or late \cite{Wang2016TSN} fusion is then performed to combine the results and make a final prediction. 
Optical flow information has also been found beneficial in few-shot video classification \cite{zhu2020lim}.
Although 3D CNNs and two-stream networks are two different family of methods, they are not mutually exclusive and can be combined together. All aforementioned methods indicate the importance of motion information in video understanding. 

Based on this observation, we believe motion information should benefit video adversarial attack. We thus propose the motion-excited sampler to generate a better prior for gradient estimation in a black-box attack setting. By incorporating motion information, our method shows superior attack performance on four benchmark datasets against two widely adopted video classification models. 

\section{Method}
\label{sec:method}

\subsection{Problem Formulation}
\label{subsec:preliminaries}
\begin{figure}[t]
\centering
\subfigure[]{
\centering
\includegraphics[width=0.56\linewidth]{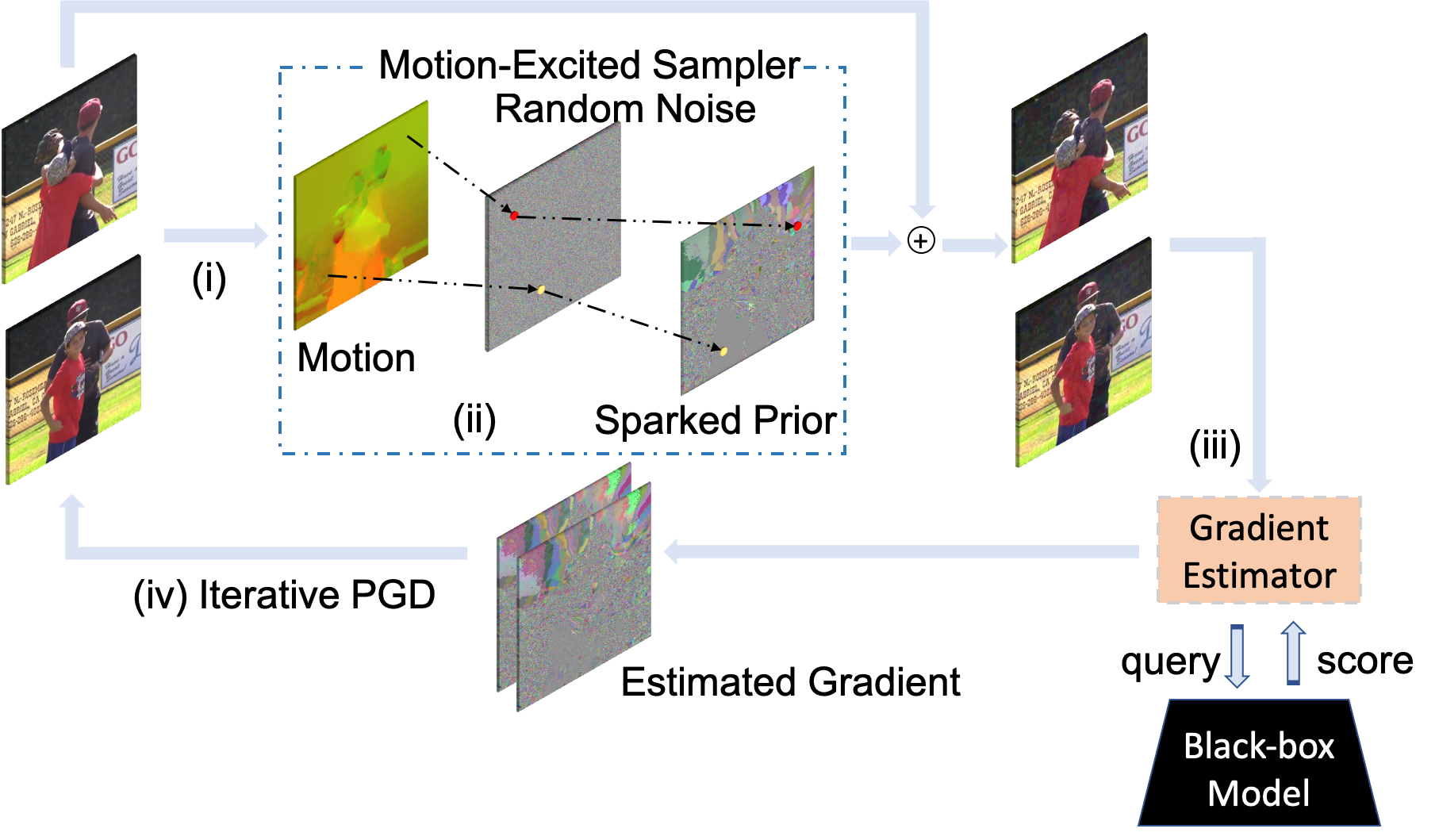}
\label{fig:framework}}
\subfigure[]{
\centering
\includegraphics[width=0.37\linewidth]{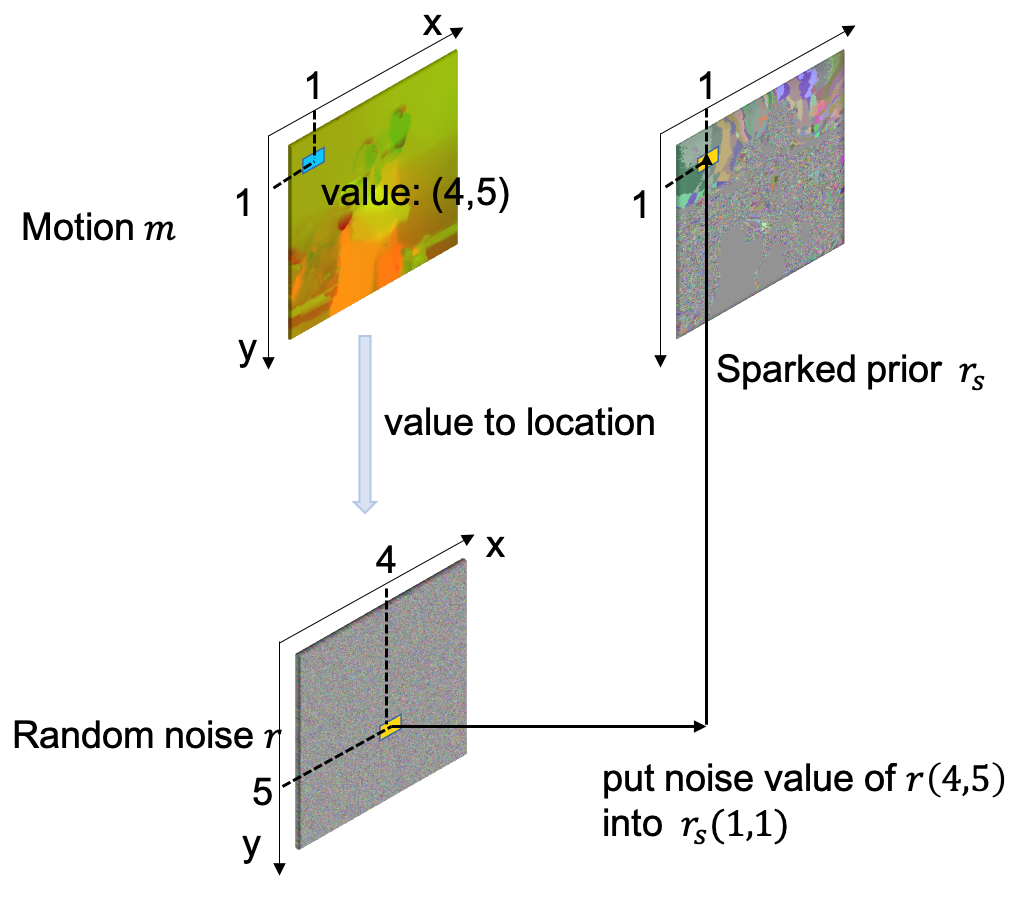}
\label{fig:simple_example}}
\caption{\textbf{(a)}: Overview of our framework for black-box video attack. i) Compute motion maps from given video frames; ii) Generate sparked prior from random noise by the proposed motion-excited sampler; iii) Estimate gradients by querying the black-box video model; iv) Use the estimated gradient to perform iterative projected gradient descent (PGD) optimization on the video. \textbf{(b)}: Illustration of Motion-Excited Sampler. }
\end{figure}

We consider a standard video action recognition task for attacking. Suppose the given videos have an underlying distribution denoted as $\mathcal{X}$, sample $\bm{x} \in \mathbb{R}^{V\times H \times W\times C}$ and its corresponding label $y\in\{1,2,...,K\}$ are a pair in $\mathcal{X}$, where $V, H, W, C$ denote the number of video frames,  height, width and channels of each frame respectively. $K$ represents the number of categories.
We denote DNN model as function $f_{\bm{\theta}}$, where $\bm{\theta}$ represents the model's parameters. The goal of a black-box adversarial attack is to find an adversarial example $\bm{x}_{adv}$ with imperceivable difference from $\bm{x}$ to fail the target model $f_{\bm{\theta}}$ through querying the target model multiple times. It can be mathematically formulated as:
\begin{equation}
\label{equ: basic problem formulation}
\begin{aligned}
    &\argmin_{\bm{x}_{adv}}L(f_{\bm{\theta}}(\bm{x}_{adv}), y)\\
    &\st~\|\bm{x}_{adv}^{i}-\bm{x}^{i}\| \leq \kappa, i = 0,1,...,V-1 \\
    &~~~~~\rm{\#queries} \leq Q
\end{aligned}
\end{equation}
Here $i$ is the video frame index, starting from $0$ to $V-1$. $\|\cdot\|$ denotes the $\ell_p$ norm that measures how much perturbation is imposed, and $\kappa$ indicates the maximum perturbations allowed. $f_{\bm{\theta}}(\bm{x})$ is the
returned logits or probability by the target model $f_{\bm{\theta}}$ when given an input video $\bm{x}$. The loss function $L(f_{\bm{\theta}}(\bm{x}_{adv}), y)$ measures the degree of certainty for the input $\bm{x}_{adv}$ maintaining true class $y$. For simplicity, we shorten the loss function as $L(\bm{x}_{adv}, y)$ in the rest of the paper since model parameters $\bm{\theta}$ remain unchanged. The goal is to minimize the certainty and successfully fool the classification model. 
The first constraint enforces high similarity between clean video $\bm{x}$ and its adversarial version $\bm{x}_{adv}$. The second constraint imposes a fixed budget $\rm{Q}$ for the number of queries used in the optimization.
Hence, the fewer queries required for adversarial video and the higher overall success rate within $\kappa$ perturbation, the better the attack method. The overview of our method is shown in Fig.~\ref{fig:framework}.

\subsection{Motion Map Generation}
\label{subsec:motionmap}
In order to incorporate motion information for video adversarial attack, we need to find an appropriate motion representation. 
There are two widely adopted motion representations in video analysis domain \cite{simonyan2014two,Wang2016TSN}, motion vector and optical flow. Both of them can reflect the pixel intensity changes between two adjacent video frames. In this work, we adopt the accumulated motion vector \cite{wu2018compressed} and the TVL1 flow \cite{chambolle2004algorithm} as the motion representation. We first describe accumulated motion vector as below.

Most modern codecs in video compression divide video into several intervals and split frames into $I$-frames (intracoded frames) and $P$-frames (predictive frames). 
Here, we denote the number of intervals as $N$ and the length of an interval as $T$. In each interval, the first frame is $I$-frame and the rest $T-1$ are $P$-frames. The accumulated motion vector is formulated as the motion vector of each $P$-frame, that can trace back to the initial $I$-frame instead of depending on previous $P$-frames.
Suppose the accumulated motion vector in frame $t$ of interval $n$ is denoted as $\bm{m}^{(t, n)}\in \mathbb{R}^{H\times W\times 2}$ and each pixel value at location $i$ in $\bm{m}^{(t, n)}$ can be computed as:
\begin{equation}
    \label{equ:(3)}
    \bm{m}_i^{(t, n)}=i-\mathcal{B}_{i}^{(t, n)}, 1\leq t\leq T-1, 0\leq n\leq N-1
\end{equation}
where $\mathcal{B}_{i}^{(t,n)}$ represents the location traced back to initial $I$-frame from $t$-th frame in interval $n$. We refer the readers to \cite{wu2018compressed} for more details.

For an interval with $T$ frames, we can obtain $T-1$ accumulated motion vectors. We only choose the last one $\bm{m}^{(T-1, n)}$ since it traces all the way back to the $I$-frame in this interval and is most motion-informative. We abbreviate $\bm{m}^{(T-1, n)}$ as $\bm{m}^{(n)}$ and thus, we have a set of $N$ accumulated motion vectors for the whole video, denoted as $\mathcal{M} = \{\bm{m}^{(0)},\bm{m}^{(1)},...,\bm{m}^{(N-1)}\}$.

Optical flow is a motion representation that is similar to motion vector with the same dimension $\mathbb{R}^{H\times W\times 2}$. 
It also contains spatial details but is not directly available in compressed videos.
Here we use TVL1 algorithm \cite{chambolle2004algorithm} to compute the flow given its wide adoption in video-related applications~\cite{Wang2016TSN,carreira2017quo,Feichtenhofer2019slowfast}.
We apply the same strategy as motion vectors to select optical flow and will also obtain $N$ flow vectors. The set of flow vectors is also denoted as $\mathcal{M}$ for simplicity.

\subsection{Motion-Excited Sampler}
\label{subsec:sampler}

In a black-box setting, random noise is normally employed in generating adversarial examples. 
However, as stated before in Fig. \ref{fig1: kinetics loss curve}(b), direct usage of random noise is not promising in video attack. To tackle this problem, we propose to involve motion information in the process of generating adversarial examples, and thus propose motion-excited sampler.

First, we define the operation of motion-excited sampler (ME-Sampler) as
\begin{equation}
    \bm{r}_{s} = \textsc{ME-Sampler}(\bm{r}, \bm{m}),
\end{equation}
where $\bm{r}\in \mathbb{R}^{V\times H\times W \times 3}$ denotes the initial random noise, motion maps $\bm{m} \in \mathbb{R}^{V\times H\times W \times 2}$ are selected with replacement from set $\mathcal{M}$ introduced in Section~\ref{subsec:motionmap}. $\bm{r}_{s} \in \mathbb{R}^{V\times H\times W \times 3}$ will be the transformed motion-aware noise, which we term as \textbf{sparked prior} afterwards.

To be specific, we use the motion-excited sampler to ``warp'' the random noise by motion. It is not just rearranging the pixels in the random noise, but constructing a completely new prior given the motion information. For simplicity, we only consider 
the operation for one frame here. It is straightforward to extend to the case of multiple frames. Without abuse of notation, we still use $\bm{r}, \bm{r}_s, \bm{m}$ for clarification in this section.

At the $i$-th location of the motion map $\bm{m}$, we denote the motion vector value as $p_i \in \mathbb{R}^2$ and its coordinate is denoted ($x_i$, $y_i$), i.e., $p_i=\bm{m}[x_i, y_i]$.
Here, $p_i = (u_i, v_i)$ has two values, which indicate the horizontal and vertical relative movements, respectively.
When computing the value of position $(x_i, y_i)$ in sparked prior $\bm{r}_s$, ($u_i, v_i$) will serve as the new coordinates for searching in original random noise. The corresponding noise value of $\bm{r}[u_i, v_i]$ will be assigned as the value in $\bm{r}_{s}[x_i, y_i]$.
Thus, we have:
\begin{align}
\label{equ: me-sampler operation}
\begin{split}
    (u_i, v_i) = \bm{m}[x_i, y_i], \\
\bm{r}_{s}[x_i, y_i] = \bm{r}[u_i, v_i].
\end{split}
\end{align}

We give a simplified example in Fig. \ref{fig:simple_example} to show how our motion-excited sampler works. To determine pixel value located in $(1, 1)$ of $\bm{r}_{s}$, we first get motion value $(4, 5)$ from motion map $\bm{m}$ at its $(1, 1)$ location. We then select pixel value located in $(4, 5)$ of $\bm{r}$ and put its value into location $(1, 1)$ of $\bm{r}_{s}$.

Generally speaking, sparked prior is still a noise map. Note that, initial noise is completely random and irrelevant to the input video. With motion-excited sampler (operation in Eq.~\ref{equ: me-sampler operation} and Fig.~\ref{fig:simple_example}), pixels with the same movements will have the same noise values in sparked prior. Then, sparked prior connects different pixels on the basis of motion map and is thus block-wised. Compared to the initial noise which is completely random, sparked prior is more informative and relevant to the video because of the incorporated motion information. It thus helps to guide the direction of estimated gradients towards attacking video models in a black-box setting and enhances the overall performance.

\begin{algorithm}[t] 
\caption{$\textsc{Grad-Est}(\bm{x}, y, \bm{g}, \rm loop)$: Estimate $\ell(\bm{g}) = -\nabla_{\bm{g}} \langle \nabla_{\bm{x}} L(\bm{x}, y), \bm{g}\rangle$.} 
\label{alg:gradient estimation} 
\begin{algorithmic}[1]
\REQUIRE video $\bm{x}$, its label $y$, number of frame of video $\bm{x}$ is $V$. initialized $\bm{g}$, interval t for sampling new motion map, $\delta$ for loss variation and $\epsilon$ for approximation.
    \IF{loop \% t = 0}
    \STATE motion map $\bm{m}_1, \bm{m}_2,...,\bm{m}_{V-1}$ are chosen from $\mathcal{M} = \{\bm{m}^{(0)},\bm{m}^{(1)},...,\bm{m}^{(N-1)}\}$ with replacement and are concatenated to be $\bm{m}$;
    \ENDIF
    \STATE $\bm{r} \leftarrow \mathcal{N}(\bm{0}, \bm{I})$;
    \STATE $\bm{r}_{s} = \textsc{ME-Sampler}(\bm{r}, \bm{m})$;
    \STATE $\bm{w}_1 = \bm{g}+\delta\bm{r}_{s}$;
    \STATE $\bm{w}_2 = \bm{g}-\delta\bm{r}_{s}$;
    \STATE $\ell(\bm{w}_1) = -\langle \nabla_{\bm{x}} L(\bm{x}, y), \bm{w}_1 \rangle \approx \frac{L(\bm{x}, y) - L(\bm{x}+\epsilon \cdot \bm{w}_1, y)}{\epsilon}$;
    \STATE $\ell(\bm{w}_2) = -\langle \nabla_{\bm{x}} L(\bm{x}, y), \bm{w}_2 \rangle \approx \frac{L(\bm{x}, y) - L(\bm{x}+\epsilon \cdot \bm{w}_2, y)}{\epsilon}$;
    \STATE $\bm{\Delta} = \frac{L( \bm{x}+\epsilon \bm{w}_2, y) - L(\bm{x}+\epsilon \bm{w}_1, y)}{\delta \epsilon}\bm{r}_{s}$;
\ENSURE  $\bm{\Delta}$.
\end{algorithmic}
\end{algorithm}

\subsection{Gradient Estimation and Optimization}
\label{subsec:optimization}
Once we have the sparked prior, we incorporate it with the input video and feed them to the black-box model to estimate the gradients.
We consider $\ell_{inf}$ noise in our paper following \cite{jiang2019black}, but our framework also applies to other norms.

Similar to \cite{ilyas2018prior}, rather than directly estimating the gradient $\nabla_{\bm{x}}L(\bm{x}, y)$ for generating adversarial video, we perform iterative updating to search. The new loss function designed for such optimization is,
\begin{equation}
\label{loss: loss for inner iteration}
    \ell(\bm{g}) = - \langle \nabla_{\bm{x}} L(\bm{x}, y), \bm{g} \rangle,
\end{equation}
$\nabla_{\bm{x}} L(\bm{x}, y)$ is the groundtruth gradient we desire and $\bm{g}$ is the gradient to be estimated. An intuitive observation from this loss function is that iterative minimization of Eq.~\ref{loss: loss for inner iteration} will drive our estimated gradient $\bm{g}$ closer to the true gradient.

We denote $\Delta$ to be the gradient $\nabla_{\bm{g}}\ell(\bm{g})$ of loss $\ell(\bm{g})$.
We perform a two-query estimation to the expectation and apply the authentic sampling to get
\begin{equation}
\label{equ: 5}
    \Delta = \frac{\ell(\bm{g}+\delta\bm{r}_{s}) - \ell(\bm{g}-\delta\bm{r}_{s})}{\delta}\bm{r}_{s},
\end{equation}
where $\delta$ is a small number adjusting the magnitude of loss variation.
By substituting Eq.~(\ref{loss: loss for inner iteration}) to Eq.~(\ref{equ: 5}), we have
\begin{equation}
\label{equ: finite difference}
    \Delta = \frac{\langle \nabla_{\bm{x}} L(\bm{x}, y),\bm{g}-\delta\bm{r}_{s}\rangle - \langle \nabla_{\bm{x}} L(\bm{x}, y),\bm{g}+\delta\bm{r}_{s} \rangle}{\delta}\bm{r}_{s}.
\end{equation}
In the context of finite difference method, we notice, given the function $L$ at a point $\bm{x}$ in the direction of vector $\bm{g}$, the directional derivative $\langle \nabla L(\bm{x}, y), \bm{g} \rangle$ can be transferred as:
\begin{equation}
\label{equ: inner gradient estimation}
    \langle \nabla_{\bm{x}} L(\bm{x}, y), \bm{g} \rangle \approx \frac{L(\bm{x}+\epsilon\bm{g}, y) - L(\bm{x},y)}{\epsilon},
\end{equation}
$\epsilon$ is a small constant for approximation. By combining Eq.~(\ref{equ: finite difference})-(\ref{equ: inner gradient estimation}), we have,
\begin{equation}
    \Delta = \frac{L( \bm{x}+\epsilon \bm{w}_2, y) - L(\bm{x}+\epsilon \bm{w}_1, y)}{\delta \epsilon}\bm{r}_{s},
\end{equation}
with $\bm{w}_1 = \bm{g}+\delta\bm{r}_{s}$ and $\bm{w}_2 = \bm{g}-\delta\bm{r}_{s}$. The resulting algorithm for generating gradient for $\bm{g}$ is shown in Algorithm~\ref{alg:gradient estimation}.

Once we have Algorithm~\ref{alg:gradient estimation}, we can use it to update estimated gradient and optimize the adversarial video. To be specific, in iteration $t$, $\Delta_t$ is returned by Algorithm~\ref{alg:gradient estimation}. We update $\bm{g}_t$ by simply applying one-step gradient descent: $\bm{g}_t = \bm{g}_{t-1} - \eta \Delta_t$, $\eta$ is a hyperparameter to update $\bm{g}_t$.
The updated $\bm{g}_t$ is the gradient we want to use for generating adversarial videos. 
Finally, we combine our estimated $\bm{g}_t$ with projection gradient descent (PGD) to translate our gradient estimation algorithm into an efficient video adversarial attack method. The detailed procedure is shown in Algorithm~\ref{alg:alg2}, in which $\argmax [f_{\bm{\theta}}(\bm{x}_t)]$ returns top predicted class label, $\textsc{Clip}(\cdot)$ constrain the updated video $\bm{x}_t$ close to the original video $\bm{x}_0$, where $\bm{x}_0 - \kappa$ is the lower bound and $\bm{x}_0+\kappa$ the upper bound. $\kappa$ is the noise constraint in Eq.~(\ref{equ: basic problem formulation}).
\begin{algorithm}[t] 
\caption{Adversarial Example Optimization for $\ell_{inf}$ norm perturbations.} 
\label{alg:alg2} 
\begin{algorithmic}[1]
\REQUIRE original video $\bm{x}$, its label $y$, learning rate $h$ for updating adversarial video.
    \STATE $\bm{x}_0 \leftarrow \bm{x}$, initially estimated $\bm{g}_0 \leftarrow \textbf{0}$, initial loop $t=1$;
    \WHILE{$\argmax [f_{\bm{\theta}}(\bm{x}_t)] = y$}
        \STATE $\bm{\Delta}_t = \textsc{Grad-Est}(\bm{x}_{t-1}, y, \bm{g}_{t-1}, t-1)$;
        \STATE $\bm{g}_t = \bm{g}_{t-1} - \eta \bm{\Delta}_t$;
        \STATE $\bm{x}_t = \bm{x}_{t-1} - h \cdot sign(\bm{g}_t)$;
        \STATE $\bm{x}_t = \textsc{Clip}(\bm{x}_t, \bm{x}_0-\kappa, \bm{x}_0+\kappa)$;
        \STATE $t = t+1$;
    \ENDWHILE
\ENSURE  $\bm{x}_t$.
\end{algorithmic}
\end{algorithm}

\subsection{Loss Function}
\label{subsec:loss}
Different from applying cross-entropy loss directly, we adopt the idea in \cite{carlini2017towards} and design a logits-based loss. Here, the logits returned from the black-box model is denoted as $l \in \mathbb{R}^{K}$, where $K$ is the number of classes. We denote the class of largest value in logits $l$ as $y$, the largest logits value is $l_y$. The final loss can be obtained as $L=\max(l_y-\max_{k \neq y}l_k, 0)$. Minimizing $L$ is expected to confuse the model with the second most confident class prediction so that our adversarial attack could succeed. 

\section{Experiments}
\label{sec:experiment}
\begin{figure*}[t]
\begin{center}
 \includegraphics[width=0.85\linewidth]{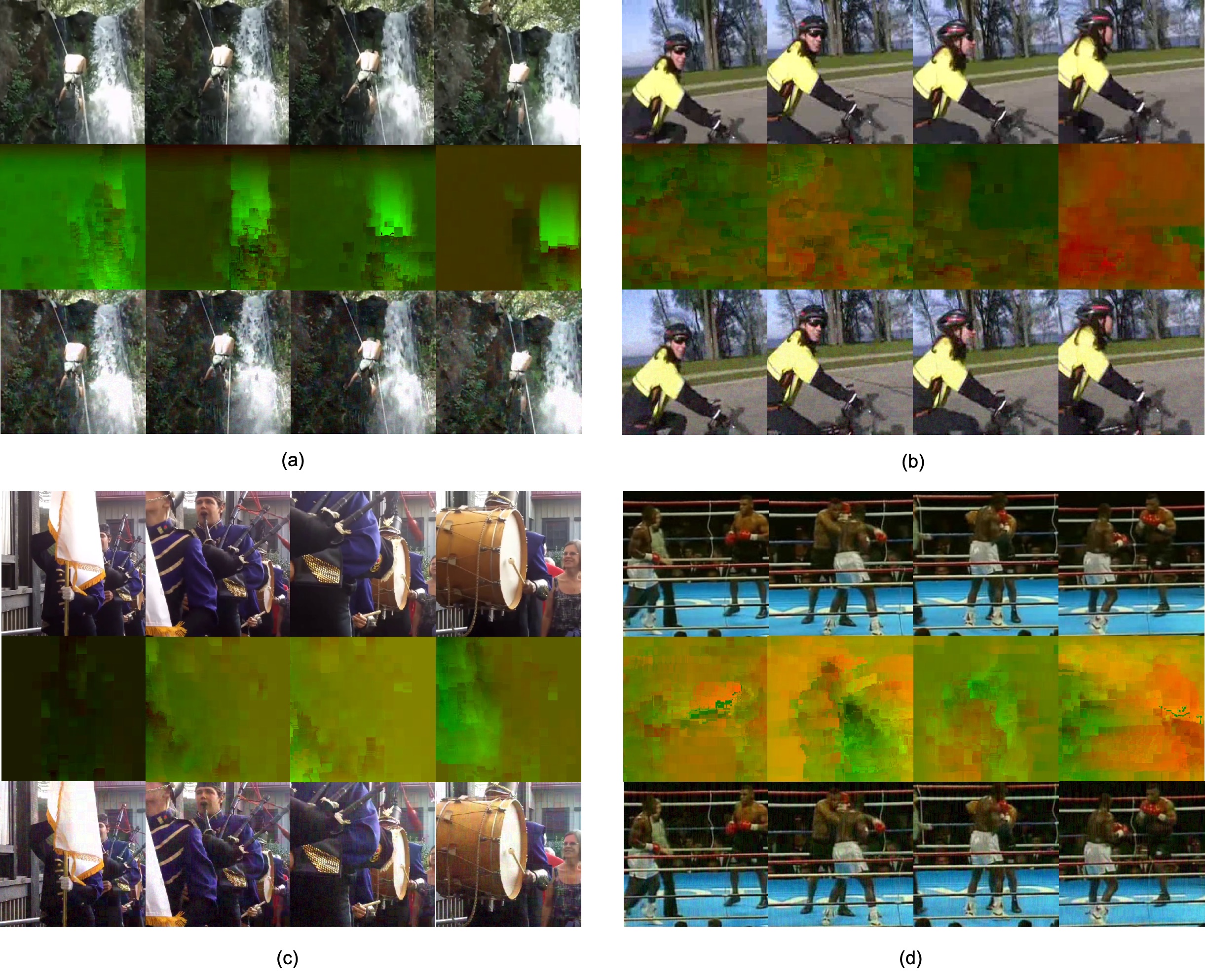}
\end{center}
   \caption{Examples of motion vectors in generating adversarial samples. In (a)-(d), the first row is the original video frame, the second row is the motion vector and the third row is generated adversarial video frame. a) Kinetics-400 on I3D: Abseling$\bm{\rightarrow}$Rock climbing; b) UCF-101 on I3D: Biking$\bm{\rightarrow}$Walking~with~dog; c) Kinetics-400 on TSN: Playing bagpipes$\bm{\rightarrow}$Playing accordion; d) UCF-101 on TSN: Punching$\bm{\rightarrow}$Lunges.}
\label{fig:visualization}
\end{figure*}

\subsection{Experimental Setting}
\noindent \textbf{Datasets.} We perform video attack on four video action recognition datasets: UCF-101~\cite{soomro2012ucf101}, HMDB-51~\cite{kuehne2011hmdb}, Kinetics-400~\cite{kay2017kinetics} and Something-Something V2~\cite{goyal2017something}. UCF-101 consists of 13,200 videos spanned over 101 action classes. HMDB-51 includes 6,766 videos in 51 classes. Kinetics-400 is a large-scale dataset which has around 300K videos in 400 classes. 
Something-Something V2 is a recent crowd-sourced video dataset on human-object interaction that needs more temporal reasoning. It contains 220,847 video clips in 174 classes. For notation simplicity, we use SthSth-V2 to represent Something-Something V2.

\noindent \textbf{Video Models.} We choose two video action recognition models, I3D \cite{kay2017kinetics} and TSN2D \cite{Wang2016TSN}, as our black-box models. For I3D training on Kinetics-400 and SthSth-V2, we train it from ImageNet initialized weights. For I3D training on UCF-101 and HMDB-51, we train it with Kinetics-400 pretrained parameters as initialization. For TSN2D training, we use ImageNet initialized weights on all four datasets.
The test accuracy of two models can be found in Table~\ref{tab:videomodels}.

\begin{table}[t]
\begin{center}
\caption{Test accuracy (\%) of the video models.}
\label{tab:videomodels}
\renewcommand\arraystretch{0.95}
\setlength{\tabcolsep}{2.5mm}
{
\begin{tabular}{ccccc}
\toprule
Model &Kinetics-400 & UCF-101 &  HMDB-51 & SthSth-V2\\
\midrule
I3D & 70.11 & 93.55 & 68.30 & 50.25\\
TSN2D & 68.87 & 86.04 & 54.83 & 35.11\\
\bottomrule
\end{tabular}}
\end{center}
\end{table}

\noindent \textbf{Attack Setting.} We perform both untargeted and targeted attack under limited queries. Untargeted attack requires the given video to be misclassified to any wrong label and targeted attack requires classifying it to a specific label. For each dataset, we 
randomly select one video from each category following the setting in \cite{jiang2019black}. All attacked videos are correctly classified by the black-box model. We impose $\ell_{inf}$ noise on video frames whose pixels are normalized to 0-1.
We constrain the maximum perturbation $\kappa = 0.03$, maximal queries Q = 60,000 for untargeted attack and $\kappa = 0.05$, Q = 200,000 for targeted attack. 
If one video is failed to attack within these constraints, we record its consumed queries as Q.

\noindent \textbf{Evaluation Metric.} 
We use the average number of queries (ANQ) required in generating effective adversarial examples and the attack success rate (SR). ANQ measures the average number of queries consumed in attacking across all videos and SR shows the overall success rate in attacking within query budget Q. A smaller ANQ and higher SR is preferred. For now, there is not a balanced metric that takes both ANQ and SR into account.

\subsection{Comparison to State-of-the-Art}
We report the effectiveness of our proposed method in Table~\ref{tab:sota}. We present the results of leveraging two kinds of motion representations: Motion Vector~(MV) and Optical Flow~(OF) in our proposed method. In comparison, we first show the attacking performance of V-BAD \cite{jiang2019black} under our video models since V-BAD is the only directly comparable method. We also extend two image attack methods~\cite{ilyas2018black,ilyas2018prior} as strong baselines to video to demonstrate the advantage of using motion information. They are denoted as E-NES and E-Bandits respectively and their attacking results are shown in Table~\ref{tab:sota}. 

Overall, our method using motion information achieves promising results on different datasets and models. On SthSth-V2 and HMDB-51, we even achieve 100\% SR. On Kinetics-400 and UCF-101, we also get over 97\% SR. The number of queries used in attacking is also encouraging. One observation worth noticing is that it only takes hundreds of queries to completely break the models on SthSth-V2.
For the rest of models, we just consume slightly more queries. To analyze this, we observe that models consuming slightly more queries often have higher recognition accuracy from Table~\ref{tab:videomodels}. From this result, we conclude that a model is likely to be more robust if its performance on the clean video is better.
\begin{table}[t]
  \begin{center}
   \caption{Untargeted attacks on SthSth-V2, HMDB-51, Kinetics-400, UCF-101. The attacked models are I3D and TSN2D. ``ME-Sampler'' denotes the results of our method. ``OF'' denotes Optical Flow. ``MV'' denotes Motion Vector.}
   \label{tab:sota}
  \renewcommand\arraystretch{0.95}
  \setlength{\tabcolsep}{2.5mm}
  {
  \begin{tabular}{cccccc}
    \toprule
    \multirow{2}{*}{Dataset~/~Model} &
    \multirow{2}{*}{Method} &
    \multicolumn{2}{c}{I3D} &
    \multicolumn{2}{c}{TSN2D}\cr
    & &ANQ & SR(\%) &ANQ & SR(\%)\\
    \midrule
    \multirow{5}{*}{SthSth-V2}
    & E-NES~\cite{ilyas2018black} &11,552 &86.96 &1,698 &99.41\\
    & E-Bandits~\cite{ilyas2018prior} &968 &100.0 &435 &99.41\\
    & V-BAD~\cite{jiang2019black}  &7,239 &97.70 &495 &100.0\\
    & ME-Sampler (OF)          &735 &98.90 &315 &100.0\\
    & ME-Sampler (MV)        &\textbf{592}  &\textbf{100.0} &\textbf{244} &\textbf{100.0}\\
    \midrule
    \multirow{5}{*}{HMDB-51}
    & E-NES~\cite{ilyas2018black} &13,237 &84.31 &19,407 &76.47\\
    & E-Bandits~\cite{ilyas2018prior} &4,549 &99.80 &4,261 &100.0\\
    & V-BAD~\cite{jiang2019black}  &5,064 &100.0 & 2,405 &100.0\\
    & ME-Sampler (OF)                &\textbf{3,306}  &\textbf{100.0} &842 &100.0\\
    & ME-Sampler (MV)             &3,915  &100.0 &\textbf{831} &\textbf{100.0}\\
    \midrule
    \multirow{5}{*}{Kinetics-400}
    & E-NES~\cite{ilyas2018black} &11,423 &89.30 &20,698 &71.93\\
    & E-Bandits~\cite{ilyas2018prior} &3,697 &99.00 &6,149 &97.50\\
    & V-BAD~\cite{jiang2019black}   &4,047 &\textbf{99.75} &2,623 & 99.75\\
    & ME-Sampler (OF)   &3,415 &99.30 &2,631 &98.80\\
    & ME-Sampler (MV)   &\textbf{2,717} &99.00 &\textbf{1,715} &\textbf{99.75} \\
    \midrule
    \multirow{5}{*}{UCF-101}
    & E-NES~\cite{ilyas2018black} &23,531 &69.23 &41,328 &34.65\\
    & E-Bandits~\cite{ilyas2018prior} &10,590 &89.10 &24,890 &66.33\\
    & V-BAD \cite{jiang2019black}  &8,819 &97.03 &17,638 & 91.09\\
    & ME-Sampler (OF)                &6,101 &96.00 &6,598 &97.00\\
    & ME-Sampler (MV)  &\textbf{4,748} &\textbf{98.02} &\textbf{5,353} &\textbf{99.00}\\
    \bottomrule
  \end{tabular}}
  \end{center}
\end{table}

In terms of using motion vector and optical flow, we find that motion vector outperforms optical flow in most cases, e.g., the number of queries used 5,353 (MV) vs 6,598 (OF) on UCF-101 against TSN2D. The reason is that motion vector always has a clearer motion region since it is computed by a small block of size $16\times16$. However, optical flow is always pixel-wisely calculated. It is not difficult to imagine that when the region used for describing is relatively larger, it is easier and more accurate to portray the overall motion. When only considering each pixel, the movement is likely to be lost in tracking and make some mistakes.

Compared to E-NES and E-Bandits, we achieve better results, either on consumed queries or success rate, i.e., when attacking a TSN2D model on UCF-101, our success rate is 99.00\%, which is much higher than 34.65\% for E-NES and 66.33\% for E-Bandits. The query 5,353 is also much smaller than 41,328 and 24,890. When compared to V-BAD, our method requires much fewer queries. For example, we save at least 1,758 queries on HMDB-51
against I3D models. Meanwhile, 
we achieve better success rate
100.0\% vs 97.70\% on SthSth-V2.

Finally, we show the visualizations of adversarial frames on Kinetics-400 and UCF-101 in Fig.~\ref{fig:visualization}. We note that the generated video has little difference from the original one but can lead to a failed prediction. More visualization can be found in the supplementary materials.
\begin{figure}[t]
\centering
\subfigure[SthSth-V2]{
   \includegraphics[width=0.47\linewidth]{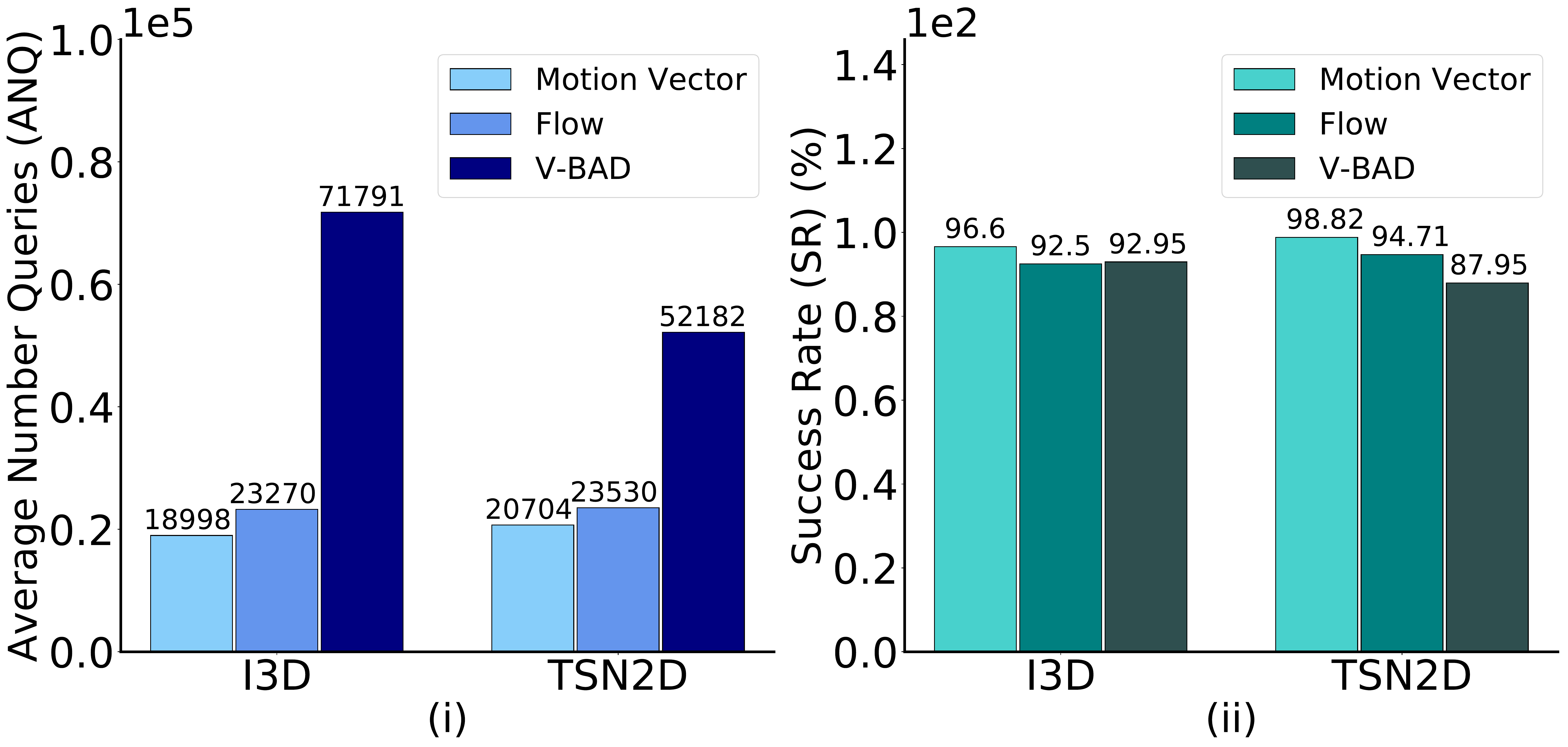}
   \label{fig: targeted attack on smth}}
\subfigure[HMDB-51]{
 \includegraphics[width=0.47\linewidth]{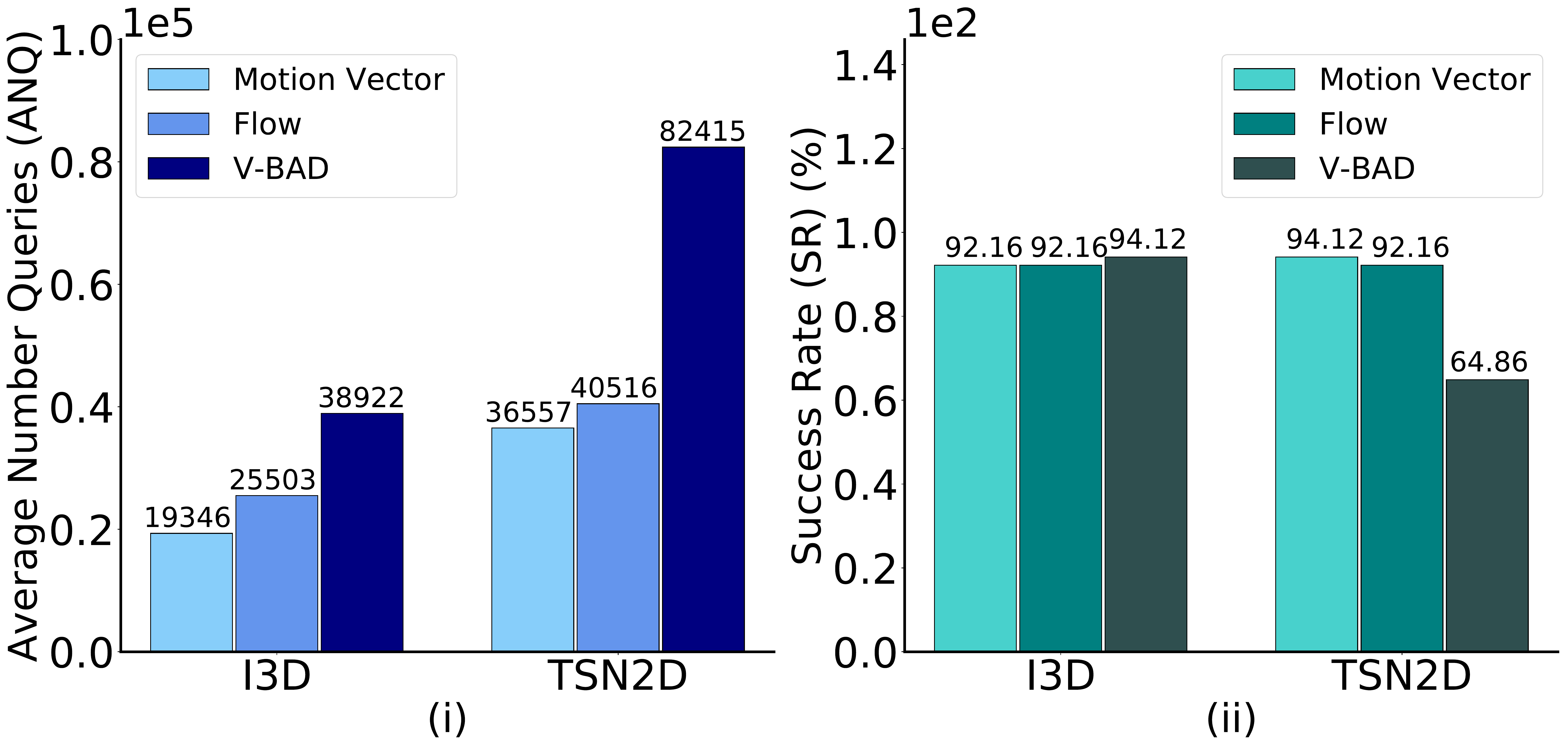}    
   \label{fig: targeted attack on hmdb}}
   \caption{\textbf{(a)}:~Comparisons of targeted attack on SthSth-V2 with V-BAD: i) Average queries consumed by I3D and TSN2D; ii) Success rate achieved by I3D and TSN2D. \textbf{(b)}:~Comparisons of targeted attack on HMDB-51 with V-BAD: i) Average queries consumed by I3D and TSN2D; ii) Success rate achieved by I3D and TSN2D.}
   \label{fig: targeted attack on smth and hmdb}
\end{figure}

\subsection{Targeted Attack}
In this section, we report the results of targeted attack on dataset HMDB-51 and SthSth-V2 in Fig.~\ref{fig: targeted attack on smth and hmdb}. For dataset SthSth-V2 from Fig.~\ref{fig: targeted attack on smth}, our method consumes less than 25,000 queries using either motion vector or optical flow. However, it costs V-BAD 71,791 against I3D model and 52,182 against TSN2D model. The success rate is about 6\% higher in TSN2D model but with much fewer queries. For dataset HMDB-51 against I3D from Fig.~\ref{fig: targeted attack on hmdb}, we also outperform V-BAD by saving more than 10,000 queries and achieve comparable success rate. For TSN2D, we only require half of the queries as V-BAD consumes but achieve a much higher success rate meanwhile, i.e., 92.16\% vs 64.86\%.

Combining with the untargeted results, we conclude that our method is more effective in generating adversarial videos than the comparing baselines.

\subsection{Ablation Study}
In this section, we first show the necessity of motion maps and then demonstrate that it is the movement pattern in the motion map that contributes to the attacking. We also study the effect of different losses. Experiments in this section are conducted on a subset of 30 randomly selected categories from UCF-101 and 80 from Kinetics-400 by following the setting in \cite{jiang2019black}.
\begin{table}[t]
\begin{center}
\caption{Compare to cases without introducing motion information.}
\label{tab:nomotion}
    \renewcommand\arraystretch{0.95}
    \setlength{\tabcolsep}{3.05mm}{
    \begin{tabular}{cccccc}
    \toprule
    \multirow{2}{*}{Dataset~/~Model} &
    \multirow{2}{*}{Method} &
    \multicolumn{2}{c}{I3D} &
    \multicolumn{2}{c}{TSN2D}\cr
    & &ANQ & SR(\%) &ANQ & SR(\%)\\
    \midrule
    \multirow{3}{*}{Kinetics-400}
    & Multi-noise &11,416 &95.00 &15,966 &89.87\\
    & One-noise &8,258 &96.25 &8,392 &96.25\\
    & Ours   &\textbf{3,089} &\textbf{100.0} &\textbf{2,494} &\textbf{100.0}\\
    \midrule
    \multirow{3}{*}{UCF-101}
    & Multi-noise &15,798 &90.00 &30,337 &70.00\\
    & One-noise &22,908 &93.33 &16,620 &90.00\\
    & Ours   &\textbf{6,876} &\textbf{100.0} &\textbf{8,399} &\textbf{100.0}\\
    \bottomrule
  \end{tabular}}
  \end{center}
\end{table}

\begin{table}[t]
  \begin{center}
  \caption{Comparison of motion map with two handcrafted maps.}
  \label{tab:motionvalue}
  \renewcommand\arraystretch{0.95}
  \setlength{\tabcolsep}{3.05mm}{
  \begin{tabular}{cccccc}
    \toprule
    \multirow{2}{*}{Dataset~/~Model} &
    \multirow{2}{*}{Method} &
    \multicolumn{2}{c}{I3D} &
    \multicolumn{2}{c}{TSN2D}\cr
    & &ANQ & SR(\%) &ANQ & SR(\%)\\
    \midrule
    \multirow{3}{*}{Kinetics-400}
    & U-Sample           &10,250 &96.25 &9,166 &96.20\\
    & S-Vaule        &8,610 &98.75 &8,429 &96.20\\
    & Our   &\textbf{3,089} &\textbf{100.0} &\textbf{2,494} &\textbf{100.0}\\
    \midrule
    \multirow{3}{*}{UCF-101}
    & U-Sample             &13,773 &93.33 &17,718 &83.33\\
    & S-Vaule     &11,471 &96.67 &17,116 &86.67\\
    & Our   &\textbf{6,876} &\textbf{100.0} &\textbf{8,399} &\textbf{100.0}\\
    \bottomrule
  \end{tabular}}
  \end{center}
\end{table}

\noindent\textbf{The necessity of motion maps.}
As mentioned in Section~\ref{sec:introduction}, we show motion is indeed important for the attack and evaluate two cases without using motion maps: 1) Multi-noise: Directly introducing random noise for each frame independently; 2) One-noise: Introducing only one random noise and replicated to all frames.
The results are in Table~\ref{tab:nomotion}. 
The results show that methods without using motion are likely to spend more queries and suffer a lower success rate. For example, on UCF-101 against I3D model, 'Multi-noise' consumes queries more than twice as our result and the success rate is 10\% lower.
Such big gaps between methods without motion and ours indicate that our designed mechanism to utilize motion maps plays an important role in directing effective gradient generation for an improved search of adversarial videos.

\noindent\textbf{Why motion maps helps?}
Here, we further replace the motion map with two handcrafted maps to reveal that the intrinsic movement pattern in a motion map matters. We show that without the correct movement pattern, the attacking performance drops significantly even using the same operation in Eq.~\ref{equ: me-sampler operation}. 

We first define a binary map $\mathcal{R}$ whose pixel values are 1 when the corresponding pixels in original motion map are nonzero, the rest pixel values are set as 0 and then define the two new maps here.
``Uniformly Sample'' (U-Sample):
    A map $\mathcal{U}$ is created whose pixel values are uniformly sampled from [0, 1] and scaled to [-50,50]. Binary map $\mathcal{R}$ and $\mathcal{U}$ are multiplied together to replace original motion map.
``Sequenced Value'' (S-Value):
    A map $\mathcal{S}$ whose pixel values are ranged in order starting from 0,1,2 from left to right and from top to bottom. Binary map $\mathcal{R}$ and $\mathcal{S}$ are multiplied together replace original motion map. 

The results are in Table~\ref{tab:motionvalue}. We first notice that `S-Value' slightly outperforms `U-Sample'. For example, on Kinetics-400 against I3D, `S-Value' saves more than 1,000 queries but gets 2\% higher success rate.
We analyze the reason to be the gradual change of pixel values in `S-Value', rather than irregular change. 
As for our method, 
on UCF-101 against TSN2D,
9,319 queries are saved and 16.77\% higher success rate is obtained when compared to `U-sample'. Through such comparison, we conclude that the movement pattern in motion map is the key factor to improve the attacking performance.
\begin{table}[t]
  \begin{center}
  \caption{Comparison of losses based on Cross-Entropy, Probability, Logits.}
  \label{tab:threeloss}
  \renewcommand\arraystretch{0.95}
  \setlength{\tabcolsep}{2.5mm}{
  \begin{tabular}{cccccc}
    \toprule
    \multirow{2}{*}{Dataset~/~Model} &
    \multirow{2}{*}{Method} &
    \multicolumn{2}{c}{I3D} &
    \multicolumn{2}{c}{TSN2D}\cr
    & &ANQ & SR(\%) &ANQ & SR(\%)\\
    \midrule
    \multirow{3}{*}{Kinetics-400}
    & Cross-Entropy              &3,452 &98.75 &2,248 &100.0\\
    & Probability                &3,089 &100.0 &2,494 &100.0\\
    & Logits                    &\textbf{2,423} &\textbf{100.0} &\textbf{1,780} &\textbf{100.0}\\
    \midrule
    \multirow{3}{*}{UCF-101}
    & Cross-Entropy             &17,362 &80.00 &17,992 &73.26\\
    & Probability               &13,217 &90.00 &14,842 &81.19\\
    & Logits                    &\textbf{6,876} &\textbf{100.0} &\textbf{6,182} &\textbf{100.0}\\
    \bottomrule
  \end{tabular}}
  \end{center}
\end{table}

\noindent\textbf{Comparison of different losses.}
We study the effect of three different losses for optimization here: cross-entropy loss, logits-based in Section~\ref{subsec:loss}. We further transfer logits $l$ to probability $p$ by Softmax and construct probability-based loss: $L\!=\!\max(\sum_{k} p_k - \max_{k \neq y}p_k, 0)$, $y$ is the class with the largest probability value. 
From the results in Table~\ref{tab:threeloss}, we conclude that logits-based loss always performs better while the other two are less effective. We also observe that Kinectics-400 is less restrictive to the selection of optimization loss, compared to UCF-101.

\section{Conclusion}
In this paper, we study the black-box adversarial attack on video models. We find that direct transfer of attack methods from image to video is less effective and hypothesize motion information in videos plays a big role in misleading video action recognition models. 
We thus propose a motion-excited sampler to generate sparked prior and obtain significantly better attack performance. 
We perform extensive ablation studies to reveal that movement pattern matters in attacking.
We hope that our work will give a new direction to study adversarial attack on video models and some insight into the
difference between videos and images.
\noindent \textbf{Acknowledgement.} 
This work is partially supported by ARC DP200100938. Hu Zhang (No. 201706340188) is partially supported by the Chinese Scholarship Council.
\bibliographystyle{splncs04}
\bibliography{eccv2020submission}

\clearpage
\appendix 
\section{Implementation details}
\label{sec:Implementation}
In terms of the accumulated motion vector, we follow the setting in \cite{wu2018compressed} and set each interval $T$ within a video to 12 frames. There is no overlap between two adjacent intervals. For each interval, we generate one accumulated motion map.
In adversarial attacking, 
we impose noises on original video frames which are normalized to [0,1]. The modified videos are processed via standard `mean' and `std' normalization
before inputting to the black-box model for gradient estimation.
Each iteration consumes three queries that two queries for $\Delta$ in Algorithm~1 and one to determine whether the updated video $\bm{x}_t$ is successful in attacking. 
In the untargeted attack setting, we set the query limit to 60,000 and the maximal iteration for updating adversarial video is 20,000.
In the targeted attack setting, we set the query limit to 200,000 and the maximal iteration is 66,667 (200,000/3).

In Algorithm~1, the interval t for sampling new motion maps is 10, $\delta$ for adjusting the magnitude of loss variation is 0.1, $\epsilon$ for approximation is 0.1. In Algorithm~2, learning rate $\eta$ for updating estimated gradient $\bm{g}_t$ is 0.1 and learning rate $h$ for updating adversarial video $\bm{x}_t$ is 0.025. 

\section{Attack transferability on flow stream}
In this section, we investigate the attack transferability of generated adversarial video on motion stream. We first train an optical flow model on original videos. We then extract new optical flows from adversarial videos generated in our attack and evaluate the performance of obtained flow model on new flows. We measure the attack transferability of our adversarial video in terms of recognition accuracy. In the experiment, we train the optical flow model on Something-Something V2, where motion information is especially important for accurate recognition.

\begin{table}[!htbp]
\caption{Transferability of adversarial video on motion stream}
\label{sample-table}
\begin{center}
\begin{tabular}{cc}
\multicolumn{1}{c}{\bf Type of flow}  &\multicolumn{1}{c}{\bf Recognition accuracy (\%)}\\
\midrule
Flow extracted from original video         &40.59 \\
Flow extracted from our adversarial video  &12.35 \\
\end{tabular}
\end{center}
\end{table}
The above table shows that using flow extracted from original videos achieves 40.59\% recognition accuracy. However, for the new flow extracted from our generated adversarial video, its recognition accuracy goes down to 12.35\%. This result clearly demonstrates that our adversarial perturbation designed for RGB stream can be transferred to attack the flow model.

\section{More visualization}
\subsubsection{Demo video} We first put together a video to showcase our generated adversarial samples. The video can be found in \href{https://drive.google.com/file/d/1v0Zfruy_gEQZlG35hdqPw3BwPdaUhN9F/view?usp=sharing}{\textit{MESampler-demo-video.mp4}}.
\subsubsection{Video and motion map visualization} 
\begin{figure*}
\begin{center}
 \includegraphics[width=0.95\linewidth]{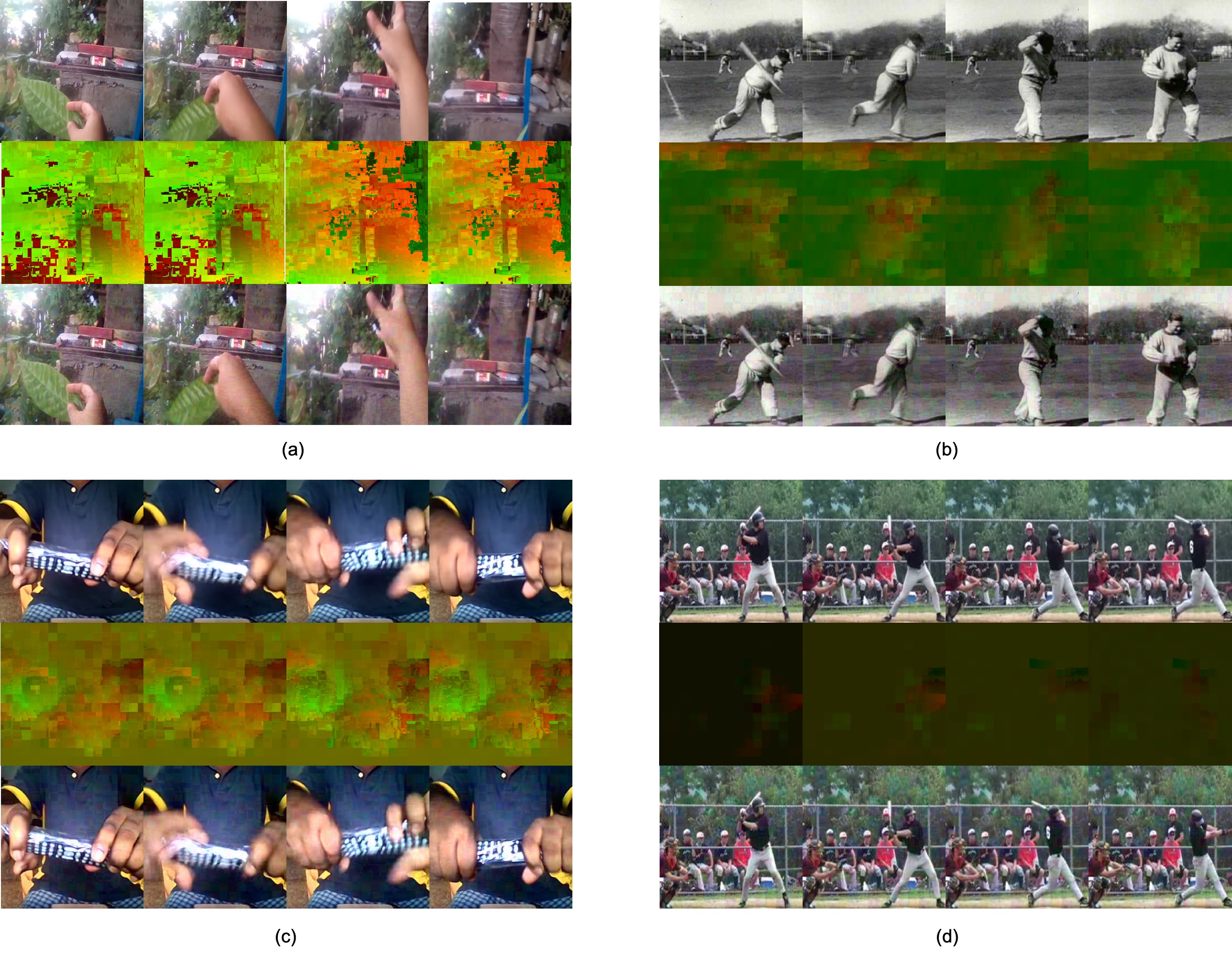}
\end{center}
   \caption{Examples of motion vectors used in attacking and the generated adversarial samples. In (a)-(d), the first row is the original video frame, the second row is the motion vector and the third row is generated adversarial video frame. a) SthSth-V2 on I3D: throwing a leaf in the air and letting it fall $\bm{\rightarrow}$ throwing tooth paste; b) HMDB-51 on I3D: throw $\bm{\rightarrow}$ fencing; c) SthSth-V2 on TSN2D: pretending or trying and failing to twist remote-control~$\bm{\rightarrow}$~pretending to open something without actually opening it; d) HMDB-51 on TSN2D: swing-baseball $\bm{\rightarrow}$ throw.}
\label{fig:smth and hmdb}
\end{figure*}

\begin{figure*}
\begin{center}
 \includegraphics[width=0.95\linewidth]{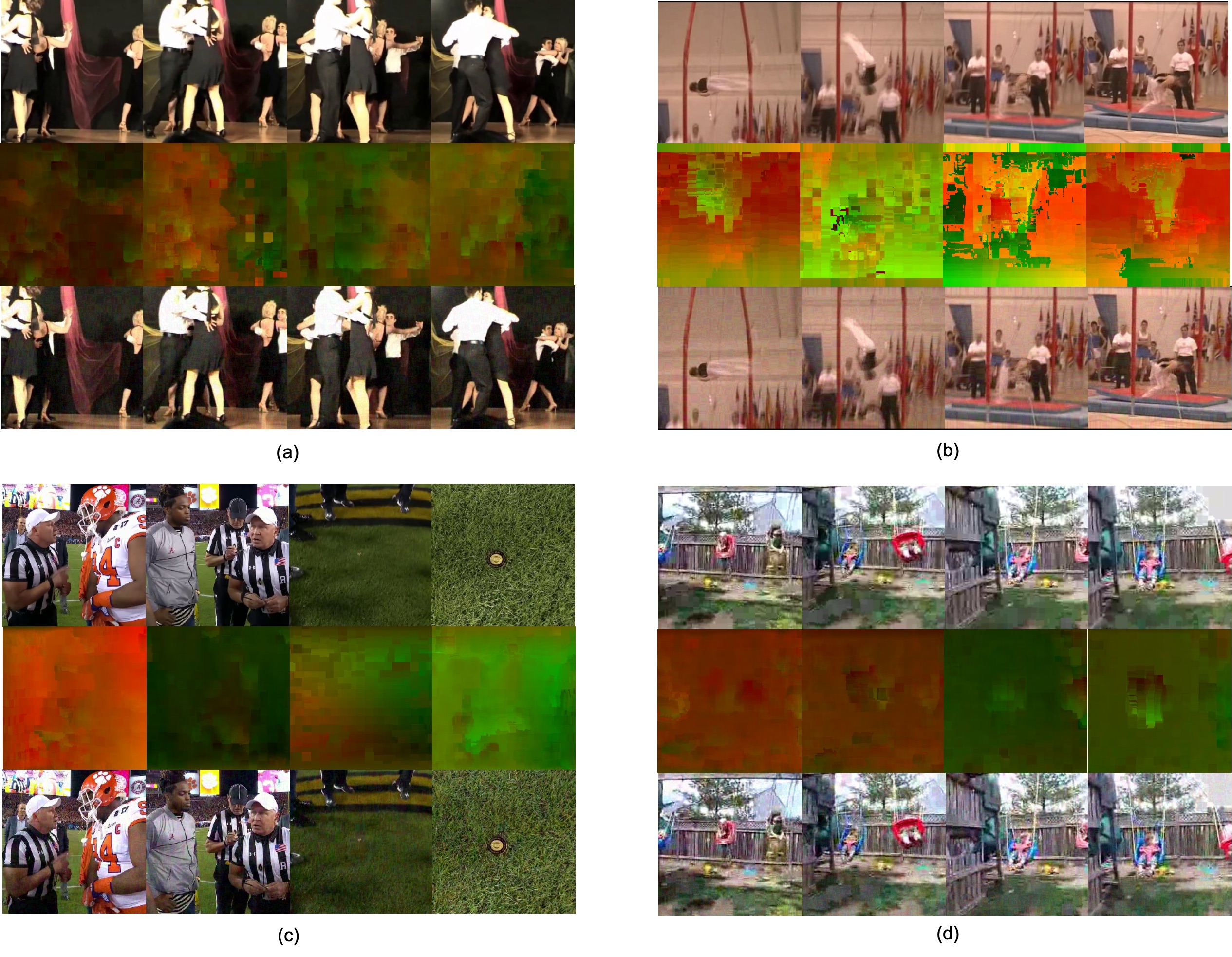}
\end{center}
   \caption{Examples of motion vectors used in attacking and generated adversarial samples. In (a)-(d), the first row is the original video frame, the second row is the motion vector and the third row is generated adversarial video frame. a) Kinetics-400 on I3D: tango dancing~$\bm{\rightarrow}$~salsa dancing; b) UCF-101 on I3D: StillRings $\bm{\rightarrow}$ PoleVault; c) Kinetics-400 on TSN2D: tossing coin $\bm{\rightarrow}$ scissors paper; d) UCF-101 on TSN2D: Swing $\bm{\rightarrow}$ TrampolineJumping.}
\label{fig:kinetics and ucf101}
\end{figure*}

We also show more results of adversarial video frames and the adopted motion vector on four datasets. Visualizations on SthSth-V2~\cite{goyal2017something} and HMDB-51~\cite{kuehne2011hmdb} are shown in Fig.~\ref{fig:smth and hmdb} and the results of Kinetics-400~\cite{kay2017kinetics} and UCF-101~\cite{soomro2012ucf101} are in Fig.~\ref{fig:kinetics and ucf101}.

From the demo video and the frame visualizations on four datasets, we can see that our generated adversarial samples can successfully fail the video classification model. Even though the generated video samples look the same as the original videos and the true labels can still be recognized by human without any difficulties.
\subsubsection{Failure cases} 
\begin{figure*}[!htbp]
\begin{center}
 \includegraphics[width=0.95\linewidth]{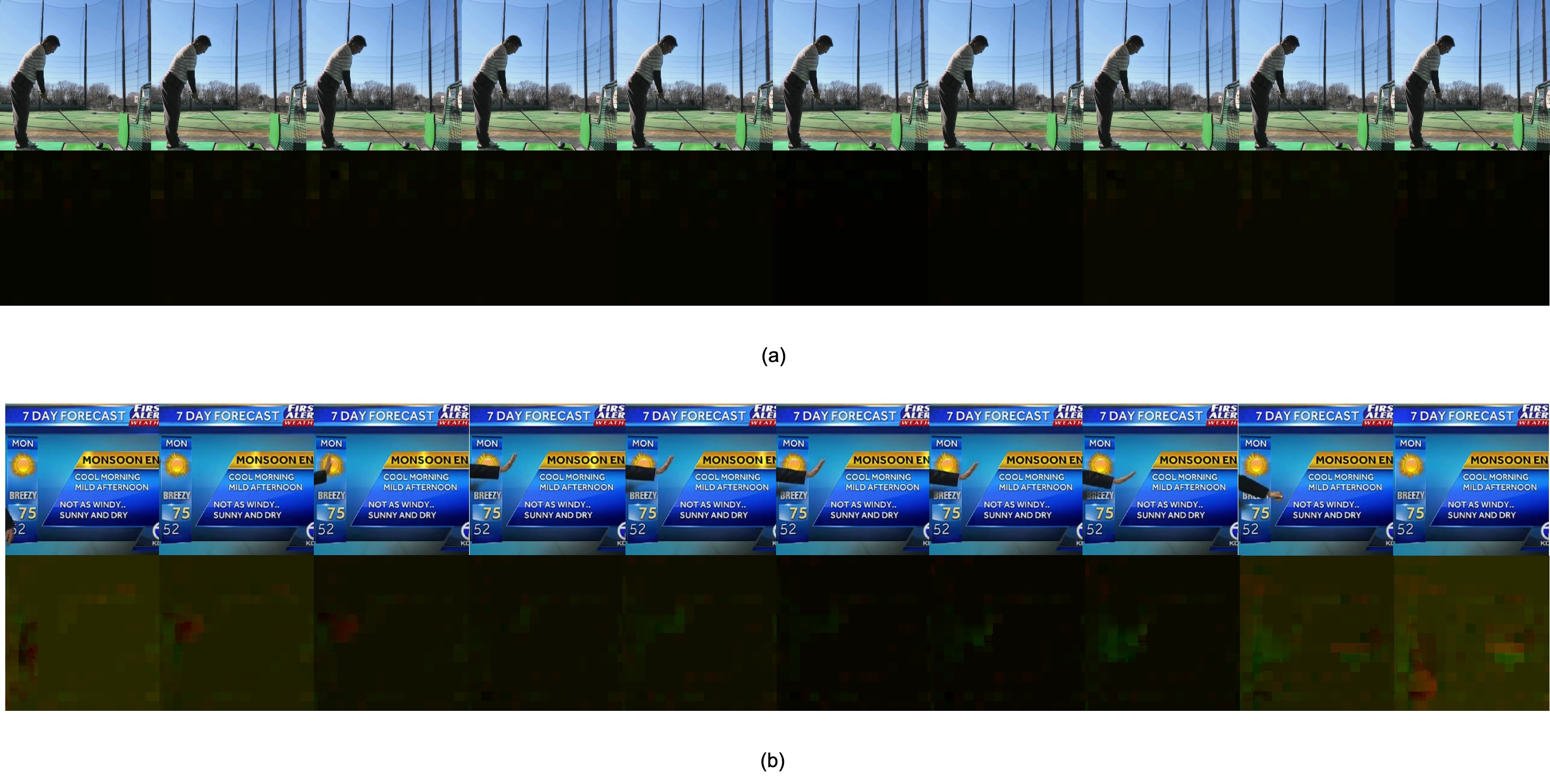}
\end{center}
   \caption{Failed samples from Kinetics-400 against I3D and TSN2D. a) Sample from class `golf driving' against I3D; b) Sample from class `presenting weather forecast' against TSN2D. The first row are the frames of original video and the second row are the motion vectors generated between frames. The movements between video frames seem to change little and the generated motion vectors  are very obscure.}
\label{fig:kinetics fail}
\end{figure*}

\begin{figure*}[!htbp]
\begin{center}
 \includegraphics[width=0.95\linewidth]{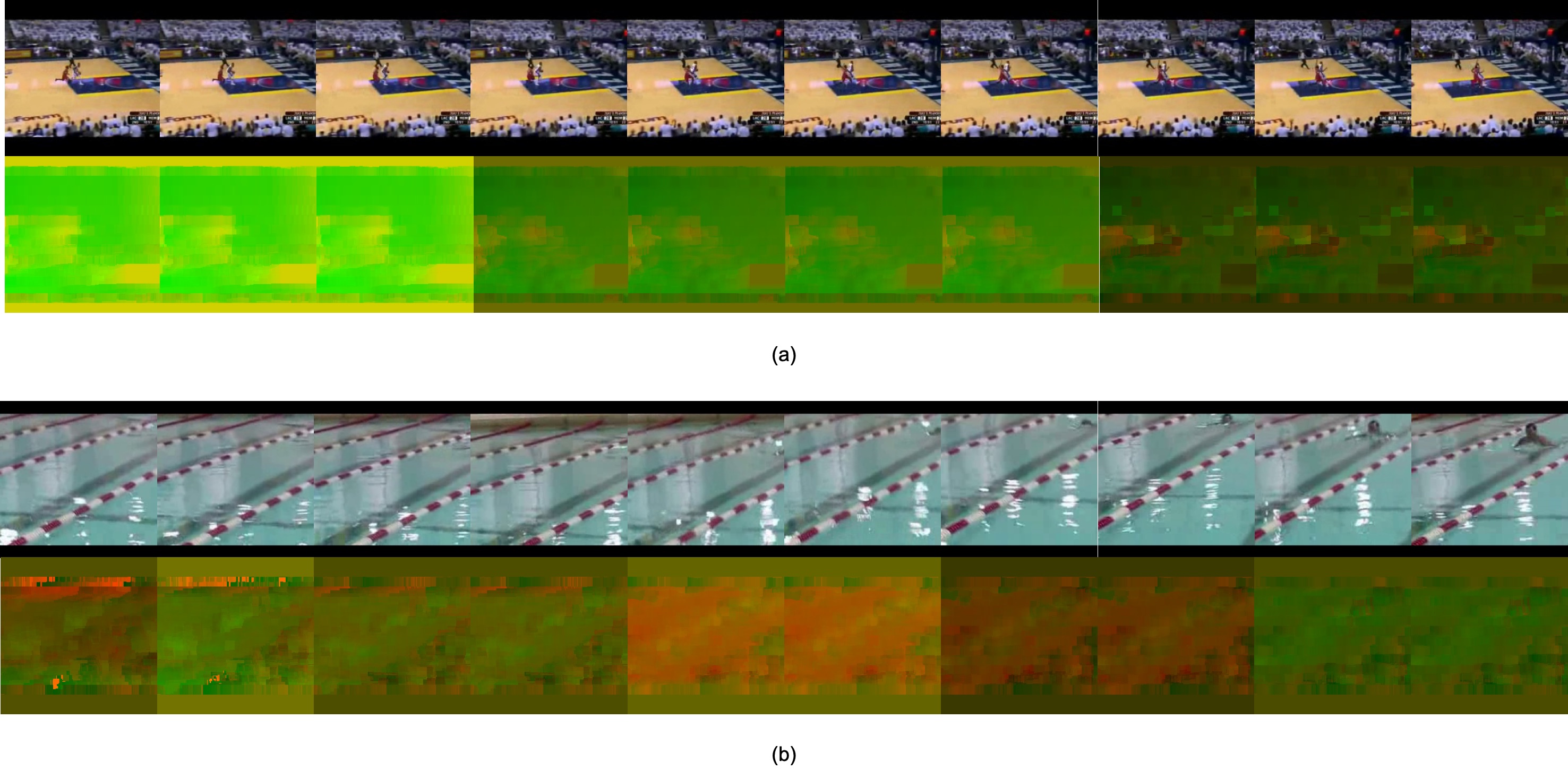}
\end{center}
   \caption{Failed samples from UCF-101 against I3D and TSN2D. a) Sample from class `BasketballDunk' against I3D; b) Sample from class `BreastStroke' against TSN2D. The first row are the frames of original video and the second row are the motion vectors generated between frames. The movement between video frames  changes little and the target object in the video is very small.}
\label{fig:ucf101 fail}
\end{figure*}

To have a better understanding of our model, we show several failure cases of our method. Examples on Kinetics-400 and UCF-101 are shown in Fig.~\ref{fig:kinetics fail} and Fig.~\ref{fig:ucf101 fail}, respectively.

There are two potential reasons behind failures in adversarial videos. The first one is about the confidence of the attacked video model. On certain videos, the model is confident about its prediction. Take the first video in Fig.~\ref{fig:kinetics fail} as an example, the black-box model outputs its true label `golf driving' with confidence 0.9999. This confidence score is so high that the perturbation posed on the video is likely to have little consequences on the final results. 
Secondly, it is about motion quality in videos. We notice that for videos that we fail to attack, their motion map is rather obscure and unrecognizable. Under such circumstances, the advantage of motion information can not be fully utilized. 
\section{Better motion information}
\begin{figure*}[!htbp]
\begin{center}
 \includegraphics[width=0.85\linewidth]{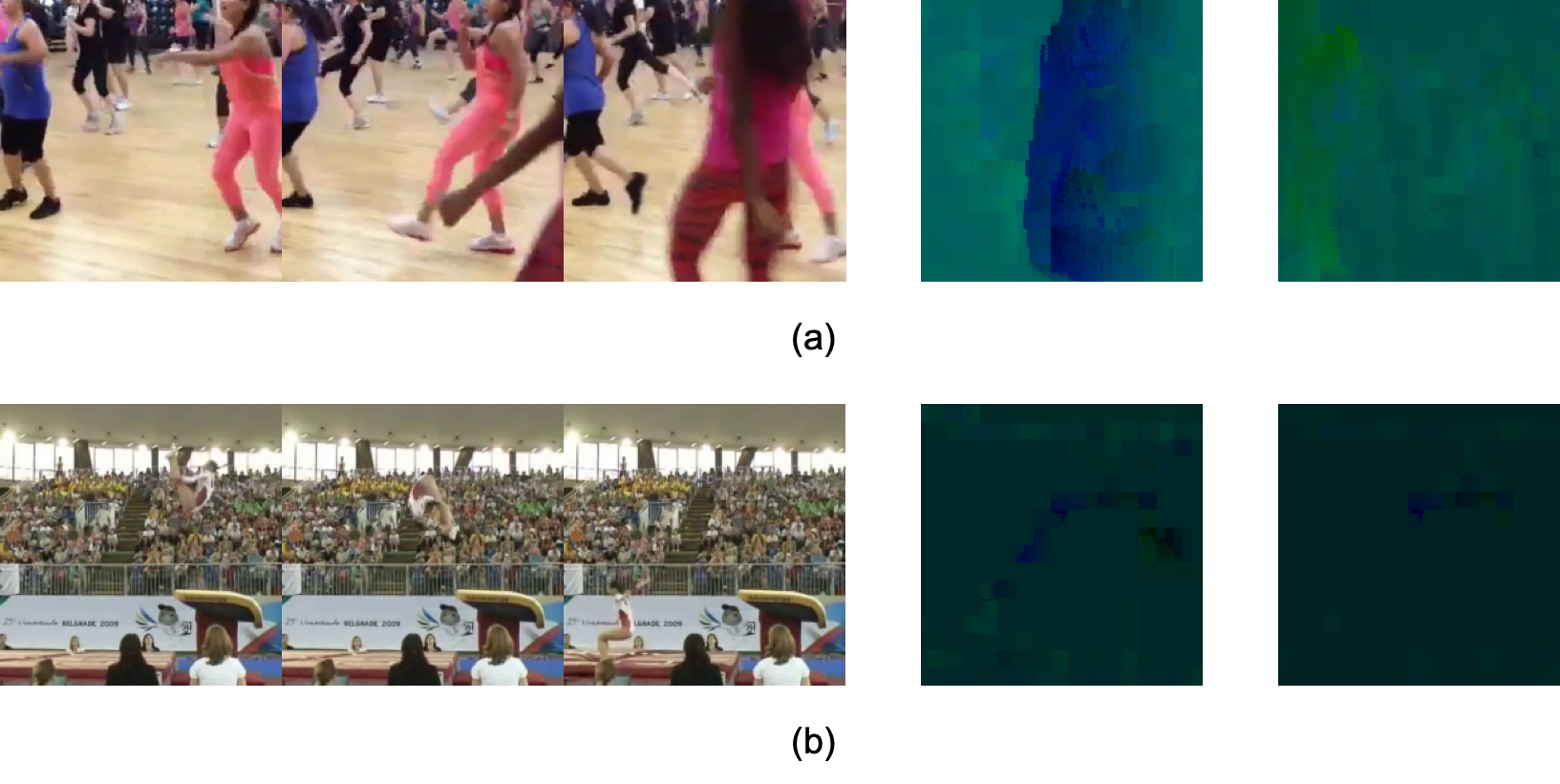}
\end{center}
   \caption{Rather than fixing the starting point and length of each interval for generating motion map, the start point and the length of interval for generating motion maps is modified according to the trajectory of the given video to get clearer and more complete description of movement. Two samples from Kinetics-400: ~\textbf{a)} Sample from class `zumba'; ~\textbf{b)} Sample from class `vault'. \textbf{Left}:~the frames of original video;~\textbf{Middle}: `improved motion map';~\textbf{Right}: original motion map in attacking. Clearly, `improved motion map' is more complete and clearer than original motion map in the right. The attacking results are also better by using the `improved motion map'.
   }
\label{fig:better motion}
\end{figure*}

Here, we justify another assumption that clearer and more completed motion maps will lead to better attack performance. Rather than fixing the starting point and length of each interval for generating motion map, the starting point and the length of interval for generating motion maps is modified according to the trajectory of given video to get a clearer and more complete description of movement. Such new map is termed as `improved motion map'. We show two samples on Kinetics-400 in Fig.~\ref{fig:better motion}, that are from class `zumba' and `vault'. The left column are the video frames, `improved motion map' and original motion map are in the middle and right respectively. Clearly, improved motion map in the middle is more consistent and clearer. For sample (a), it saves more than 10,000 queries by applying `improved motion map' instead of original one. For sample (b), 30,000 queries are also saved by using the `improved motion map'. 

However, it is still difficult to automatically determine the starting point and the interval length to generate much clearer maps. We will leave the study as the future work.


\end{document}